\documentclass[letterpaper]{article} 
\usepackage[preprint]{aaai2027}  
\usepackage[hyphens]{url}  
\usepackage{graphicx} 
\usepackage{natbib}  
\usepackage{caption} 
\usepackage{algorithm}
\usepackage{algorithmic}
\usepackage{multirow, multicol}
\usepackage{amsmath}
\usepackage{amssymb}
\usepackage{mathtools}
\usepackage{pifont}

\newcommand{\pub}[1]{{\color{gray}{\tiny{[{#1}]}}}}
\usepackage{pifont}
\usepackage{makecell}

\usepackage{newfloat}
\usepackage{listings}
\DeclareCaptionStyle{ruled}{labelfont=normalfont,labelsep=colon,strut=off} 

\floatstyle{ruled}
\newfloat{listing}{tb}{lst}{}
\floatname{listing}{Listing}

\usepackage{booktabs}

\title{ExPhy: A Benchmark for Explicit Physical Property Learning in\\ Multi-Object Trajectory Forecasting}
\author{
    Rui Wang\textsuperscript{\rm 1},
    Yeteng Wu\textsuperscript{\rm 1},
    Xianlin Zhang\textsuperscript{\rm 2},
    Mengshi Qi\corresponding\textsuperscript{\rm 1}
}

\affiliations{
    \textsuperscript{\rm 1}State Key Laboratory of Networking and Switching Technology,
    Beijing University of Posts and Telecommunications\\
    \textsuperscript{\rm 2}School of Digital Media \& Design Art,
    Beijing University of Posts and Telecommunications\\
    Beijing, China\\
    wr@bupt.edu.cn,
    wuyeteng@bupt.edu.cn,
    zxlin@bupt.edu.cn,
    qms@bupt.edu.cn
}

\begin{document}

\maketitle

\begin{abstract}
Understanding object dynamics requires not only predicting future trajectories
but also examining whether a model captures the physical properties that govern motion. However, existing benchmarks rarely expose object-level physical properties
as explicit evaluation targets alongside trajectory forecasting. To address this gap, we introduce \emph{ExPhy}, a multi-object trajectory forecasting benchmark containing 24,000 simulated physical scenes with explicit
object-level labels for mass, friction, and restitution. ExPhy provides observed
and future trajectories together with an in-distribution (ID) split and two out-of-distribution (OOD) splits over physical parameters (OOD-Parameter) and initial states (OOD-Initial)
for jointly evaluating trajectory forecasting and physical
property estimation. We further instantiate
\textsc{PhyODE}, a physics-guided model with an explicit property
interface that estimates physical
properties from observed trajectories and uses them for differentiable future
rollout. On the long-horizon OOD-Initial setting, \textsc{PhyODE} reduces ADE
and FDE by 33.1\% and 31.0\%, respectively, compared with the strongest
baseline. Zero-shot evaluation on ComPhy further assesses cross-benchmark transfer.
Property-level analyses reveal that accurate trajectory forecasting does not
necessarily imply accurate recovery of the underlying physical properties.
Code and data are available at~\url{https://github.com/Zest86/ExPhy}.
\end{abstract}

\section{Introduction}

Understanding object dynamics is a central problem in physical reasoning, as
predicting future motion from observed trajectories requires accounting for
the physical properties that shape how objects move and
interact~\cite{battaglia2016interaction,watters2017visual}. Humans exhibit intuitive physical reasoning from early
development, such as anticipating motion, collision outcomes, and material
responses~\cite{davis2008physical,spelke2007core,wu2024physical,qi2026explainable,qi2026chain}. In dynamic
multi-object interactions, properties such as mass, friction, and restitution
govern inertial response, tangential contact behavior, and collision rebound,
respectively. Existing evaluations of object-centric
physical dynamics commonly focus on future outcomes or predicted trajectories,
while estimation of the underlying object-level properties is often assessed
separately or not at all. This distinction matters because low trajectory
error does not necessarily imply accurate physical property estimation. It therefore motivates evaluating trajectory forecasting alongside explicit object-level property learning.

\begin{figure}[!t]
	\centering
	\includegraphics[width=\linewidth]{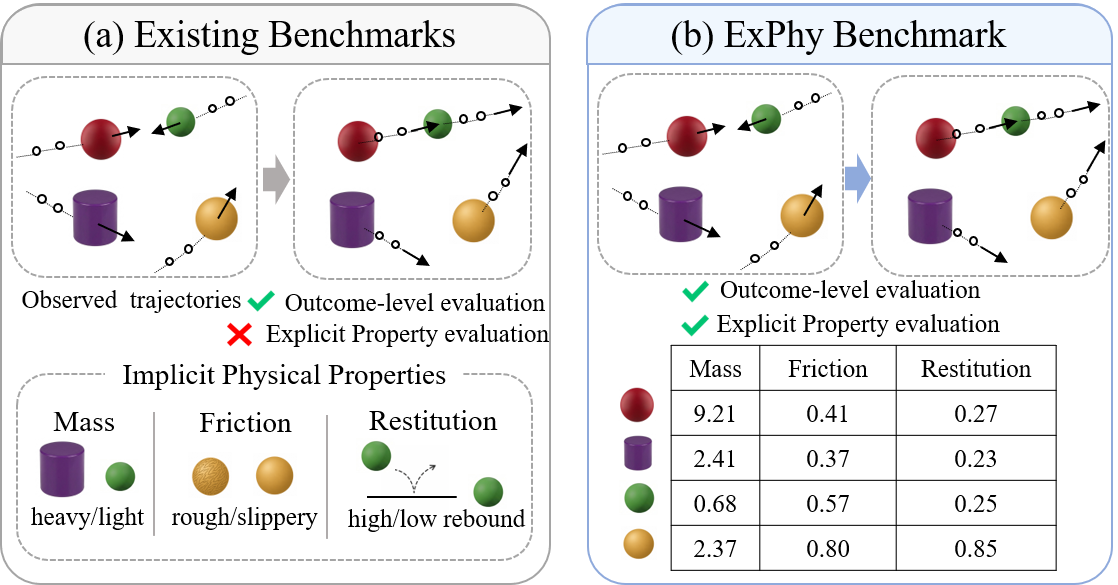}
    \caption{Comparison between existing benchmarks and the proposed ExPhy
    benchmark. (a) Existing benchmarks primarily supervise future outcomes, while
    object-level physical properties are often represented implicitly or indirectly
    and are unavailable as dedicated evaluation targets. (b) ExPhy provides
    explicit continuous-valued labels for mass, friction, and restitution, enabling
    joint evaluation of trajectory forecasting and physical property estimation.}
	\label{fig:teaser}  
    \vspace{-5mm}
\end{figure}

\begin{table}[t]
    \centering
    \caption{
    Comparison of representative physical reasoning benchmarks. $\circ$ denotes property-dependent evaluation without direct property targets.
    Obj. denotes direct object-level property evaluation; Cont. denotes
    continuous parameter regression; Traj. denotes trajectory-level
    forecasting; and OOD denotes a controlled distribution-shift protocol.
    }
    
    \label{tab:benchmark_comparison}
    \resizebox{\columnwidth}{!}{%
        \begin{tabular}{lcccc}
            \toprule
            \multirow{2}{*}{Dataset}
            & \multicolumn{2}{c}{Phys. Prop. Eval.}
            & \multirow{2}{*}{Traj. Eval.}
            & \multirow{2}{*}{OOD Eval.} \\
            \cmidrule(lr){2-3}
            & Obj-level & Cont. & & \\
            \midrule

            CLEVRER \pub{ICLR20}
            & $\times$ & $\times$ & $\times$ & $\times$ \\

            Super-CLEVR \pub{CVPR23}
            & $\times$ & $\times$ & $\times$ & $\checkmark$ \\


            Physion++ \pub{NeurIPS23}
            & $\circ$ & $\times$ & $\times$ & $\times$ \\

            ComPhy \pub{TPAMI25}
            & $\checkmark$ & $\times$ & $\times$ & $\times$ \\

            PhysBench \pub{ICLR25}
            & $\checkmark$ & $\times$ & $\times$ & $\times$ \\

            PhysInOne \pub{CVPR26}
            & $\checkmark$ & $\checkmark$ & $\times$ & $\times$ \\


            \midrule
            \textbf{ExPhy (Ours)}
            & $\checkmark$ & $\checkmark$
            & $\checkmark$ & $\checkmark$ \\
            \bottomrule
        \end{tabular}%
    }
\end{table}

Despite recent progress in physical reasoning and dynamics prediction, existing
benchmarks still largely emphasize future-outcome prediction or task-specific reasoning, including future-state prediction, physical question answering, and event plausibility
judgment~\cite{clevrer,super-clevrer-3d,bear2021physion,chen2025compositional,qi2025robust,wang2026vto}. As illustrated in
Fig.~\ref{fig:teaser} and summarized in
Table~\ref{tab:benchmark_comparison}, object-level physical factors are often
represented implicitly or indirectly, rather than exposed as dedicated
object-level evaluation targets. To address this gap, we introduce \emph{ExPhy}, a benchmark for joint trajectory and physical-property evaluation in multi-object trajectory forecasting.
ExPhy comprises 24k simulated dynamic scenes with observed and future
trajectories and explicit object-level labels for mass, friction, and
restitution. Together with dedicated in-distribution (ID) and out-of-distribution (OOD) splits, it enables unified
evaluation of trajectory forecasting, physical property estimation, and
generalization under controlled distribution shifts.

With ExPhy, we distinguish two complementary questions: whether a model
forecasts future trajectories accurately and whether its estimated physical
properties agree with the underlying simulator parameters. Existing trajectory forecasting
models typically learn future states directly from observed trajectories through
graph interactions, Transformers, or latent dynamics
models~\cite{li2020visual,han2022learning,huang2022equivariant,
liu2023segno,wen2022social,NRMF,fu2025moflow,qi2025action}. Although effective under
ADE/FDE, their internal variables need not correspond to physically meaningful
properties. To instantiate the ExPhy evaluation, we develop \textsc{PhyODE}, a
physics-guided hybrid model that estimates object-level physical properties
from observed trajectories and uses them for differentiable future rollout.
Property labels are used as training supervision but are never provided as
inference inputs. We evaluate \textsc{PhyODE} under ID and OOD settings, conduct zero-shot
transfer to ComPhy, and analyze object-level property estimation. Together,
these experiments expose the distinction between accurate trajectory
forecasting and accurate physical property estimation.

The main contributions are summarized as follows:

(1) We introduce \textit{ExPhy}, a benchmark comprising 24k multi-object scenes with
explicit object-level labels for mass, friction, and restitution, together with ID,
OOD-Parameter, and OOD-Initial evaluation protocols.

(2) We develop \textsc{PhyODE}, a physics-guided hybrid model with an explicit property interface that estimates
object-level physical properties from observed trajectories and uses them
for differentiable future rollout.

(3) Extensive experiments demonstrate competitive ID/OOD and cross-benchmark
forecasting, and reveal that trajectory accuracy and physical property
accuracy are related but distinct evaluation dimensions.

\section{Related Work}

\paragraph{Physical Reasoning Benchmarks.}
Existing benchmarks evaluate complementary aspects of physical understanding.
CLEVRER~\cite{clevrer} focuses on causal and future-event reasoning, while
Super-CLEVR~\cite{super-clevrer} introduces controlled domain shifts for
compositional visual reasoning. Physion and Physion++~\cite{bear2021physion,physion++}
primarily evaluate future-contact or outcome prediction, which require inferring
latent mechanical properties. ComPhy~\cite{chen2025compositional} directly evaluates
object-level mass and charge through categorical targets, whereas
PhysBench~\cite{chow_physbench_2025} assesses broader physical properties through
multiple-choice questions. PhysInOne~\cite{zhou_physinone_2026} further supports
continuous parameter estimation and physics-based resimulation. Overall, prior
benchmarks address property reasoning, future prediction, and distribution shifts,
but largely through separate protocols. ExPhy instead unifies direct continuous
object-level property evaluation, future trajectory forecasting, and controlled
extrapolation over physical parameters and initial states.

\paragraph{Trajectory and Dynamics Forecasting.}
Trajectory forecasting methods predict future motion using recurrent
networks~\cite{alahi2016social,Qi_2020_CVPR},
Transformers~\cite{zhang2024decouple,zhou2025siam}, interaction models, or
generative frameworks~\cite{fu2025moflow,NRMF}. Despite strong ADE/FDE performance,
their representations need not correspond to physically meaningful object
properties. Physics-informed methods introduce analytical dynamics, differentiable
simulators, or structural constraints~\cite{tang_intrinsic,
Differentiable_Physics_Simulation,xu2024learning,wang2025pitn}, while Neural ODE
approaches~\cite{huang2020learning,wen2022social,luo2023hope,yuan2024egode}
model continuous-time evolution. However, these models often use entangled latent
states without direct supervision or evaluation of named physical properties.

\paragraph{Physical Property Learning.}
Related studies infer hidden physical properties from visual observations or
interactions. Latent dynamics approaches~\cite{battaglia2016interaction,
watters2017visual,zhu2024latent} encode physical information without requiring
interpretable variables. Direct estimators predict quantities such as mass,
material, or interaction parameters, but may rely on appearance
cues~\cite{wu2015galileo,image2mass}, semantic priors~\cite{zhai2024physical},
multi-view observations~\cite{li2023pac}, or foundation
models~\cite{zhan2025inferring}. Property estimation is also commonly evaluated
separately from future trajectory forecasting. PhyODE instead estimates mass,
friction, and restitution from observed trajectories and integrates them into
differentiable rollout, enabling trajectory and property evaluation within a
single model.

\section{ExPhy benchmark}

We introduce \emph{ExPhy}, a benchmark for jointly evaluating multi-object
trajectory forecasting and explicit physical property estimation. ExPhy
provides observed and future trajectories together with object-level mass,
friction, and restitution annotations, as well as controlled distribution
shifts over physical parameters and initial states.

\textbf{Problem Formulation.} Each ExPhy instance contains the observed trajectories of $N$ interacting
objects. For each object $i\in\{1,\ldots,N\}$, let
$\mathbf{x}_i^t=[x_i^t,y_i^t,z_i^t]^\top\in\mathbb{R}^3$
denote its 3D position at time step $t$, and let
$
\mathbf{X}_{i,\mathrm{obs}}
=
(\mathbf{x}_i^1,\ldots,\mathbf{x}_i^{T_{\mathrm{obs}}})
$
denote its observed trajectory. We collect all object trajectories as
$\mathcal{X}_{\mathrm{obs}}
=(\mathbf{X}_{1,\mathrm{obs}},\ldots,
\mathbf{X}_{N,\mathrm{obs}})
\in\mathbb{R}^{N\times T_{\mathrm{obs}}\times 3}$.
Given $\mathcal{X}_{\mathrm{obs}}$, the primary task is to predict the future
trajectories
$\widehat{\mathcal{X}}_{\mathrm{pred}}
\in\mathbb{R}^{N\times T_{\mathrm{pred}}\times 3}$
over the following $T_{\mathrm{pred}}$ steps. ExPhy additionally provides an
explicit property vector
$\mathbf{p}_i=[m_i,\mu_i,e_i]^\top$
for each object, corresponding to mass, friction, and restitution. These
labels are not provided as inference inputs, but support property-supervised
training and property-level evaluation.

\begin{figure}[t]
	\centering
	\includegraphics[width=\linewidth]{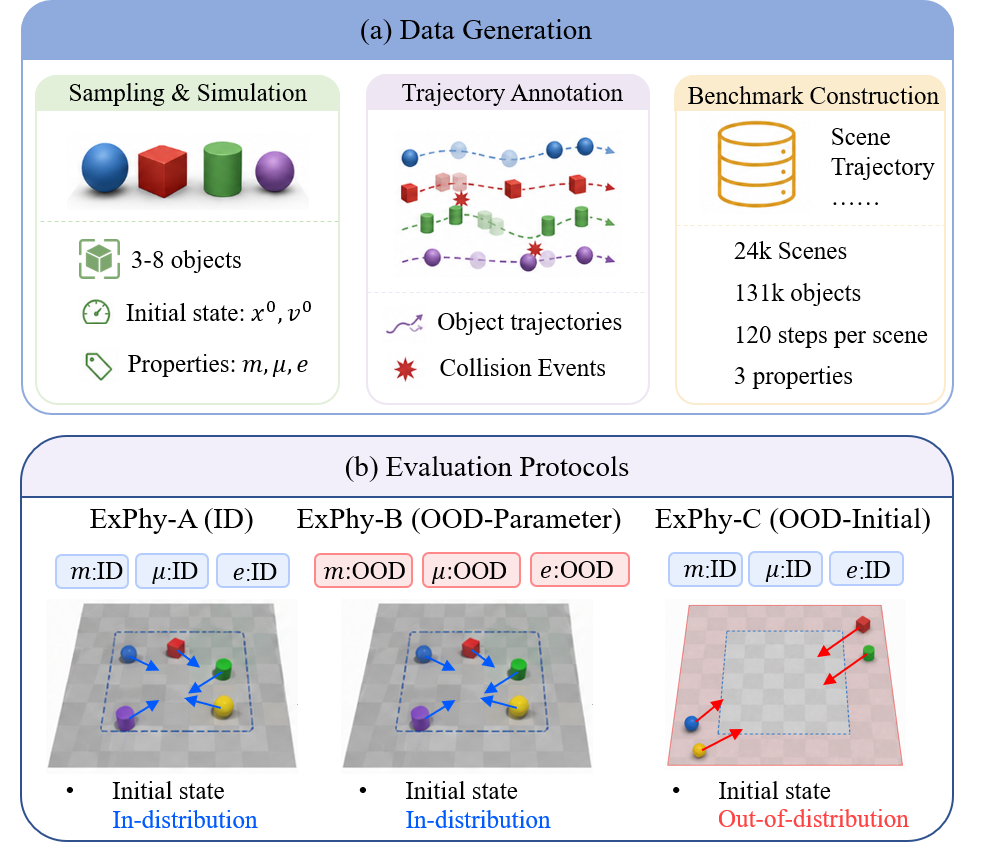}
    \caption{Overview of the ExPhy benchmark. (a) Benchmark construction and annotation pipeline. (b) ID, OOD-Parameter, and OOD-Initial evaluation protocols. Blue and red denote ID and OOD variables, respectively; the blue dashed box and red-shaded outer region mark their location ranges, while arrows depict initial velocities.}
	\label{fig: benchmark}
    \vspace{-2mm}
\end{figure}

\subsection{Dataset Construction}

ExPhy is constructed using the PyBullet physics engine~\cite{pybullet} to
generate controllable multi-object rigid-body interactions. Each scene
contains 3--8 objects with diverse geometric shapes, including cubes,
cylinders, and spheres. Their initial states and physical properties are
sampled from predefined distributions to produce diverse motion patterns and
collision events. ExPhy contains 24k dynamic scenes. ExPhy-A comprises 20k scenes, split into
16k/2k/2k training, validation, and test sets, while ExPhy-B and ExPhy-C each
contain 2k held-out OOD test scenes. Object trajectories and object-level mass,
friction, and restitution labels are recorded directly from the simulator.
Figure~\ref{fig: benchmark} illustrates the construction and evaluation
protocols, and Table~\ref{tab:exphy_sampling_ranges} summarizes the sampling
ranges.

\begin{table}[t]
    \centering
    \caption{Sampling ranges and supports for the ExPhy splits. For friction
    and restitution, one of the two listed intervals is selected uniformly at
    random before sampling within it. ExPhy-B shifts the physical-property
    distributions, whereas ExPhy-C shifts the initial-state distributions.
    ``Same'' denotes the corresponding ExPhy-A setting.}
    \label{tab:exphy_sampling_ranges}
    \footnotesize
    \setlength{\tabcolsep}{3.2pt}
    \renewcommand{\arraystretch}{1.08}

    \begin{tabular*}{\columnwidth}
        {@{\extracolsep{\fill}}lccc@{}}
        \toprule
        \textbf{Variable}
        & \textbf{ExPhy-A}
        & \textbf{ExPhy-B}
        & \textbf{ExPhy-C} \\
        \midrule

        \multicolumn{4}{@{}l}{\textit{Physical properties}} \\
        \addlinespace[1pt]

        \textbf{Mass} $m$
        & $[0.1,10]$
        & $[10.01,15]$
        & Same \\

        \cmidrule(lr){1-4}

        \multirow{2}{*}{\textbf{Friction} $\mu$}
        & $[0.35,0.60]$
        & $[0.25,0.34]$
        & \multirow{2}{*}{Same} \\
        & $[0.70,0.95]$
        & $[0.96,1.00]$
        & \\

        \cmidrule(lr){1-4}

        \multirow{2}{*}{\textbf{Restitution} $e$}
        & $[0.15,0.40]$
        & $[0.05,0.14]$
        & \multirow{2}{*}{Same} \\
        & $[0.55,0.85]$
        & $[0.86,0.95]$
        & \\

        \midrule
        \multicolumn{4}{@{}l}{\textit{Initial state}} \\
        \addlinespace[1pt]

        \textbf{Location} $\mathbf{x}_{i,xy}^{1}$
        & $[-7,7]^2$
        & Same
        & $[-10,10]^2 \setminus [-7,7]^2$ \\
        
        \textbf{Velocity} $\mathbf{v}^1_{i,xy}$ 
        & $[-3,3]^2$
        & Same
        & $\bigl([-5,-3]\cup[3,5]\bigr)^2$  \\

        \bottomrule
    \end{tabular*}
    \vspace{-3mm}
\end{table}

\begin{figure*}[t]
	\centering
	\includegraphics[width=\linewidth]{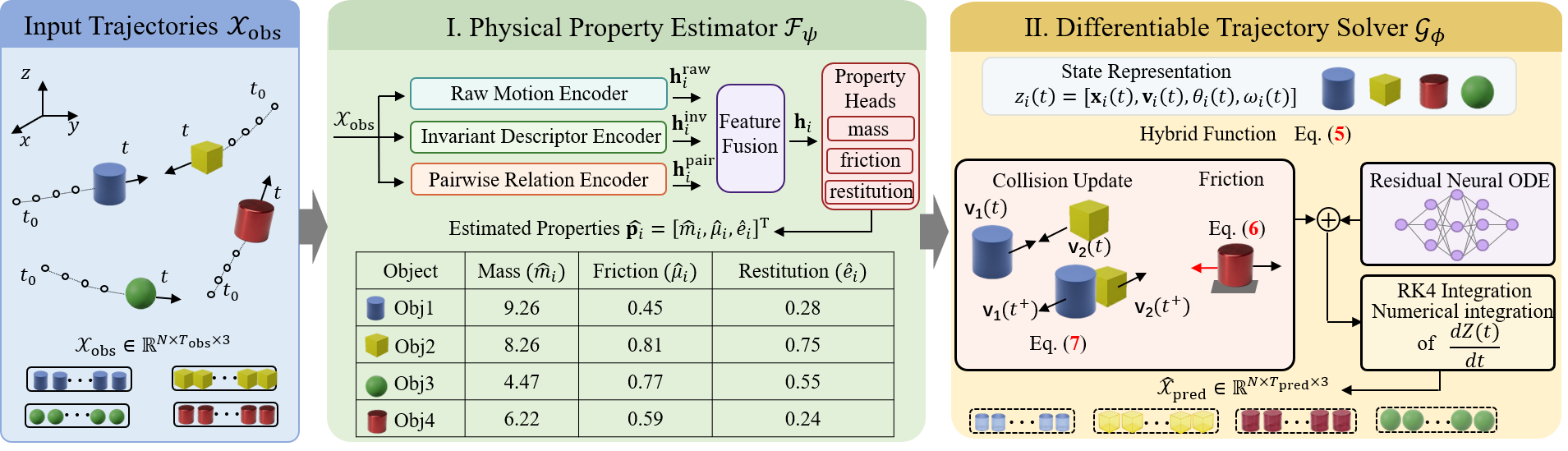}
    \caption{Overview of \textsc{PhyODE}. The physical property estimator $\mathcal{F}_{\psi}$ combines raw motion, invariant trajectory, and pairwise relation features to estimate object-level mass, friction, and
    restitution. Conditioned on these properties, the differentiable trajectory solver $\mathcal{G}_{\phi}$ combines frictional dissipation and discrete collision updates with a residual Neural ODE, and uses RK4 integration to forecast future trajectories.}
	\label{fig: model overview}
    \vspace{-4mm}
\end{figure*}

\subsection{Explicit Physical Property Labels}
For each object $i$, ExPhy provides an explicit physical property vector
$\mathbf{p}_i=[m_i,\mu_i,e_i]^\top$, comprising mass, friction, and
restitution. These labels correspond to the simulator mass, lateral friction coefficient, and
restitution parameters, which affect inertial response, tangential contact
behavior, and collision rebound, respectively. The property labels are recorded directly from the simulation configuration
and remain constant for each object throughout a scene. Because these
properties are not directly observable from a single 3D position, their
estimation relies on temporal motion and inter-object interaction cues. The
explicit annotations support supervised property learning and direct
object-level property evaluation. The sampling ranges across ExPhy-A/B/C are
summarized in Table~\ref{tab:exphy_sampling_ranges}.

\subsection{Evaluation Protocols}
\label{sec:ood}

To evaluate generalization beyond the training distribution, ExPhy provides
three complementary splits. \textbf{ExPhy-A (In-Distribution)} follows the same
physical-property and spatial distributions across training, validation, and
test sets, and measures standard interpolation performance.
\textbf{ExPhy-B (OOD-Parameter)} evaluates extrapolation to unseen physical
properties by sampling mass, friction, and restitution outside the training
ranges while retaining the same spatial distribution.
\textbf{ExPhy-C (OOD-Initial)} evaluates extrapolation to unseen initial states by shifting both the initial-location and initial-velocity distributions while preserving the ExPhy-A physical-property ranges. All models are trained on ExPhy-A and directly evaluated on ExPhy-B/C without
fine-tuning. We further define three observation-prediction horizons: Short
($10$--$10$), Mid ($20$--$40$), and Long ($30$--$60$), covering increasingly
challenging forecasting durations.

\section{Methodology}

\subsection{Overview}

As shown in Figure~\ref{fig: model overview}, \textsc{PhyODE} is a
physics-guided hybrid model that couples an explicit property estimator with
a differentiable trajectory solver. Given $\mathcal{X}_{\mathrm{obs}}$,
$\mathcal{F}_{\psi}$ estimates object-level mass, friction, and restitution,
which condition $\mathcal{G}_{\phi}$ to combine physics-based dynamics with a
residual Neural ODE and produce $\hat{\mathcal{X}}_{\mathrm{pred}}$.
The model is trained end-to-end with trajectory and property supervision,
while inference uses only the observed trajectories without ground-truth
property labels.

\subsection{Explicit Physical Property Estimator $\mathcal{F}_\psi$}
Given the observed scene trajectories
$\mathcal{X}_{\mathrm{obs}}
=
\{\mathbf{X}_{i,\mathrm{obs}}\}_{i=1}^N
\in
\mathbb{R}^{N\times T_{\mathrm{obs}}\times 3}$,
the estimator extracts complementary cues through three encoding branches.
The raw motion encoder captures coordinate-level temporal evolution from
positions and their first- and second-order differences:
\begin{equation}
    \mathbf{h}^{\mathrm{raw}}_i
    =
    f_{\mathrm{raw}}
    \left(
    \left[
    \mathbf{X}_{i,\mathrm{obs}};
    \Delta\mathbf{X}_{i,\mathrm{obs}};
    \Delta^2\mathbf{X}_{i,\mathrm{obs}}
    \right]
    \right),
\end{equation}
where $\Delta\mathbf{X}_{i,\mathrm{obs}}$ and
$\Delta^2\mathbf{X}_{i,\mathrm{obs}}$ represent the finite-difference
velocities and accelerations, respectively.

Complementarily, the invariant descriptor encoder summarizes trajectory
geometry independently of absolute coordinates:
\begin{equation}
    \mathbf{h}^{\mathrm{inv}}_i
    =
    f_{\mathrm{inv}}
    \left(
    \Phi_{\mathrm{inv}}
    (\mathbf{X}_{i,\mathrm{obs}})
    \right),
\end{equation}
where $\Phi_{\mathrm{inv}}$ concatenates relative temporal changes and
motion-magnitude statistics, with the complete descriptor definition provided
in the supplementary material.

The pairwise relation encoder captures interaction-dependent physical cues.
For each object pair $(i,j)$, relation features are constructed from relative
positions, relative velocities, and pairwise distances, and then aggregated
using learned attention weights:
\begin{equation}
    \mathbf{r}_{ij}
    =
    \Phi_{\mathrm{pair}}
    \left(
    \mathbf{X}_{i,\mathrm{obs}},
    \mathbf{X}_{j,\mathrm{obs}}
    \right), \quad
    \mathbf{h}^{\mathrm{pair}}_i=
    \sum_{j\neq i}
    \alpha_{ij}
    f_{\mathrm{pair}}(\mathbf{r}_{ij}).
\end{equation}
This aggregation is permutation equivariant with respect to object ordering
and allows the estimator to identify physical cues revealed through
inter-object interactions.

The three branch representations are fused into an object-level latent
feature:
\begin{equation}
    \mathbf{h}_i
    =
    f_{\mathrm{fusion}}
    \left(
    \left[
    \mathbf{h}^{\mathrm{raw}}_i;
    \mathbf{h}^{\mathrm{inv}}_i;
    \mathbf{h}^{\mathrm{pair}}_i
    \right]
    \right).
\end{equation}
The fused feature is decoded by three property-specific regression heads to
produce
$\hat{\mathbf{p}}_i
=
[\hat{m}_i,\hat{\mu}_i,\hat{e}_i]^\top
=
\mathcal{F}_{\psi}(\mathcal{X}_{\mathrm{obs}})_i$.
We use a Softplus output for positive mass prediction and sigmoid outputs to
constrain friction and restitution to $[0,1]$.
\subsection{Physics-Based Hybrid Trajectory Solver $\mathcal{G}_{\phi}$}

\paragraph{Hybrid dynamics formulation.}
Given the estimated object-level physical properties
$\hat{\mathbf{p}}_i=[\hat{m}_i,\hat{\mu}_i,\hat{e}_i]^\top$,
the trajectory solver rolls out future object states in continuous time.
We use $\mathbf{x}_i(t)$ to denote the continuous-time counterpart of the
discretely observed coordinates $\mathbf{x}_i^t$. For each object $i$, we define an eight-dimensional state
$\mathbf{z}_i(t)=[\mathbf{x}_i(t),\mathbf{v}_i(t),
\theta_i(t),\omega_i(t)]$, where
$\mathbf{x}_i(t)=[x_i(t),y_i(t),z_i(t)]$ contains the 3D coordinates,
$\mathbf{v}_i(t)=[v_{i,x}(t),v_{i,y}(t), v_{i,z}(t)]$ denotes the corresponding
velocity, and $\theta_i(t)$ and $\omega_i(t)$ represent scalar planar
orientation and angular velocity, respectively.

Let
$\mathbf{Z}(t)=[\mathbf{z}_1(t),\ldots,\mathbf{z}_N(t)]$
and
$\hat{\mathbf{P}}=[\hat{\mathbf{p}}_1,\ldots,\hat{\mathbf{p}}_N]$.
The hybrid dynamics combine a property-conditioned physics branch with a
learnable residual vector field:
\begin{equation}
    \frac{d\mathbf{Z}(t)}{dt}
    =
    f_{\mathrm{DPE}}
    \bigl(\mathbf{Z}(t),\hat{\mathbf{P}}\bigr)
    +
    \lambda_{\mathrm{res}}
    f_{\phi}^{\mathrm{res}}
    \bigl(\mathbf{Z}(t),t\bigr).
    \label{eq:hybrid_dynamics}
\end{equation}
Here, $f_{\mathrm{DPE}}$ models the continuous dynamics under
kinetic friction, while $f_{\phi}^{\mathrm{res}}$ provides learned corrections
to the translational and angular derivatives. The DPE additionally handles
collision detection and impulse-based state updates during rollout.

\paragraph{Physics-based dynamics and numerical rollout.}
During each collision-free interval, the continuous component of the DPE
advances the translational and angular states according to
\begin{equation}
\begin{aligned}
    \frac{d\mathbf{x}_i(t)}{dt}
    &=
    \mathbf{v}_i(t), &
    \frac{d\mathbf{v}_i(t)}{dt}
    &=
    -\hat{\mu}_i g
    \frac{\mathbf{v}_{i,\parallel}(t)}
    {\|\mathbf{v}_{i,\parallel}(t)\|_2+\epsilon}, \\
    \frac{d\theta_i(t)}{dt}
    &=
    \omega_i(t), &
    \frac{d\omega_i(t)}{dt}
    &=
    0,
\end{aligned}
\end{equation}
where $g$ is the gravitational acceleration,
$\mathbf{v}_{i,\parallel}(t)$ is the velocity tangent to the supporting
surface, and $\epsilon$ ensures numerical stability. This defines the
continuous DPE step, followed by impulse-based collision updates to the
linear and angular velocities.

After continuous integration, the DPE applies an impulse-based update to each
detected collision. Superscripts $-$ and $+$ denote the states immediately
before and after the impulse update, respectively. For an approaching pair
$(i,j)$, let $\mathbf{n}_{ij}$ and $\mathbf{t}_{ij}$ denote the contact normal
and tangent, and define the corresponding relative velocities as
$v_{ij}^{n}=(\mathbf{v}_i^{-}-\mathbf{v}_j^{-})^\top\mathbf{n}_{ij}$ and
$v_{ij}^{t}=(\mathbf{v}_i^{-}-\mathbf{v}_j^{-})^\top\mathbf{t}_{ij}$.
The normal and tangential impulse components are
$J_{ij}^{n}
=-(1+\hat{e}_{ij})v_{ij}^{n}/
(\hat{m}_i^{-1}+\hat{m}_j^{-1})$
and
$J_{ij}^{t}
=-\hat{\mu}_{ij}|J_{ij}^{n}|
\operatorname{sign}(v_{ij}^{t})$,
respectively. Defining the total impulse as
$\mathbf{J}_{ij}
=J_{ij}^{n}\mathbf{n}_{ij}
+J_{ij}^{t}\mathbf{t}_{ij}$,
the linear and angular velocities are updated by
\begin{equation}
\begin{aligned}
    \mathbf{v}_i^{+}
    &=
    \mathbf{v}_i^{-}
    +\hat{m}_i^{-1}\mathbf{J}_{ij},
    &
    \mathbf{v}_j^{+}
    &=
    \mathbf{v}_j^{-}
    -\hat{m}_j^{-1}\mathbf{J}_{ij}, \\
    \omega_i^{+}
    &=
    \omega_i^{-}+\frac{J_{ij}^{t}r_i}{I_i},
    &
    \omega_j^{+}
    &=
    \omega_j^{-}-\frac{J_{ij}^{t}r_j}{I_j}.
\end{aligned}
\label{eq:collision_update}
\end{equation}
Here, the DPE uses the symmetric pairwise coefficients
$\hat{e}_{ij}=(\hat{e}_i+\hat{e}_j)/2$ and
$\hat{\mu}_{ij}=(\hat{\mu}_i+\hat{\mu}_j)/2$.
The effective planar object scale $r_i$ is obtained from the observed state
and defines the corresponding effective moment of inertia
$I_i=\frac{1}{2}\hat{m}_i r_i^2$.
The instantaneous impulse update changes the linear and angular velocities
while leaving $\mathbf{x}_i$ and $\theta_i$ unchanged.

Let $\mathbf{Z}_k=\mathbf{Z}(t_0+k\Delta t)$ denote the rollout state at the
$k$-th prediction step. At each step, the DPE computes the
property-conditioned friction and collision responses from the current state.
These physics-based dynamics are combined with the residual vector field and
integrated using RK4:
\begin{equation}
    \mathbf{Z}_{k+1}
    =
    \operatorname{RK4}
    \left(
    \mathbf{Z}_k,
    f_{\mathrm{DPE}}
    +
    \lambda_{\mathrm{res}}f_{\phi}^{\mathrm{res}},
    \Delta t
    \right).
\end{equation}
After $T_{\mathrm{pred}}$ steps, the predicted trajectories are
\begin{equation}
    \hat{\mathcal{X}}_{\mathrm{pred}}
    =
    \left[
    \Pi_{\mathbf{x}}(\mathbf{Z}_1),
    \ldots,
    \Pi_{\mathbf{x}}(\mathbf{Z}_{T_{\mathrm{pred}}})
    \right]
    \in
    \mathbb{R}^{N\times T_{\mathrm{pred}}\times 3},
\end{equation}
where $\Pi_{\mathbf{x}}(\mathbf{Z}_k)\in\mathbb{R}^{N\times3}$ extracts the
3D coordinates of all objects at prediction step $k$.

\subsection{Training Objective}

\textsc{PhyODE} is trained end-to-end using both trajectory supervision and
physical-property supervision. The trajectory loss is defined over all objects
and future time steps:
\begin{equation}
    \mathcal{L}_{\mathrm{traj}}
    =
    \frac{1}{N T_{\mathrm{pred}}}
    \sum_{i=1}^{N}
    \sum_{t=1}^{T_{\mathrm{pred}}}
    \left\|
    \hat{\mathbf{x}}_i^t-\mathbf{x}_i^t
    \right\|_2^2,
\end{equation}
where $N$ denotes the number of objects in the scene.

For physical-property supervision, prediction errors are normalized using
fixed property-specific scales $r_p$, shared across training and evaluation.
The property loss is then given by
\begin{equation}
    \mathcal{L}_{\mathrm{prop}}
    =
    \frac{1}{3N}
    \sum_{i=1}^{N}
    \sum_{p\in\{m,\mu,e\}}
    \operatorname{SmoothL1}
    \left(
    \frac{\hat{p}_i-p_i}{r_p}
    \right).
\end{equation}

The overall training objective is
\begin{equation}
    \mathcal{L}_{\mathrm{total}}
    =
    \mathcal{L}_{\mathrm{traj}}
    +
    \lambda_{\mathrm{prop}}
    \mathcal{L}_{\mathrm{prop}},
\end{equation}
where $\lambda_{\mathrm{prop}}$ balances trajectory forecasting and physical
property estimation. 

\begin{table*}[ht]
	\centering
	\caption{Quantitative comparison of trajectory forecasting error (ADE/FDE $\downarrow$) on ExPhy-A, ExPhy-B and ExPhy-C. Lower is better. The prediction horizons are explicitly defined based on observation-prediction steps ($T_{\text{obs}}$-$T_{\text{pred}}$): Short (10-10), Mid (20-40), and Long (30-60). $\dagger$ indicates trajectory-only adaptations of physical reasoning baselines, where visual/perceptual frontends are replaced with trajectory encoders while preserving their original reasoning mechanisms. The baselines are grouped according to their primary inductive biases. \textbf{Bold} and \underline{underlined} indicate the best and second-best results, respectively.}
	\label{tab:main_results_merged}
	\resizebox{\textwidth}{!}{
		\begin{tabular}{l ccc ccc ccc}
			\toprule
			\multirow{2}{*}{\textbf{Methods}} 
			& \multicolumn{3}{c}{\textbf{ExPhy-A} (In-Distribution)} 
			& \multicolumn{3}{c}{\textbf{ExPhy-B} (OOD-Parameter)} 
			& \multicolumn{3}{c}{\textbf{ExPhy-C} (OOD-Initial)} \\
			\cmidrule(lr){2-4} \cmidrule(lr){5-7} \cmidrule(lr){8-10}
			& Short & Mid & Long 
			& Short & Mid & Long 
			& Short & Mid & Long \\
			\midrule
			
			\multicolumn{10}{l}{\textit{Physical reasoning baselines}} \\
			VRDP$^\dagger$~\pub{NeurIPS21} 
			& \underline{0.04}/\underline{0.08} 
			& 0.28/0.58 
			& 0.41/0.83 
			& \textbf{0.04}/\underline{0.08} 
			& 0.29/0.61 
			& 0.45/0.92 
			& \underline{0.12}/0.23 
			& 0.93/1.93 
			& 2.13/4.23 \\
			
	       	PHYCINE$^\dagger$~\pub{CVPR23} 
			& \underline{0.04}/\underline{0.08} 
			& 0.34/0.66 
			& 0.46/0.90 
			& \textbf{0.04}/\underline{0.08} 
			& 0.36/0.70 
			& 0.51/1.01 
			& \underline{0.12}/\underline{0.21} 
			& 1.12/2.15 
			& 1.95/3.86 \\

            PCR$^\dagger$~\pub{TPAMI25} 
			& 0.05/0.10 
			& 0.28/0.57 
			& 0.48/0.94 
			& 0.05/0.10 
			& \underline{0.28}/0.58 
			& 0.52/1.04 
			& 0.14/0.28 
			& \underline{0.67}/\underline{1.47} 
			& 1.58/3.27 \\
            
			\midrule
			\addlinespace[2pt]
			\multicolumn{10}{l}{\textit{Geometric dynamics baselines}} \\
			
            PAINET~\pub{ICLR26} 
            & 0.05/0.10 
            & \underline{0.27}/0.57 
            & \underline{0.40}/0.81 
            & 0.06/0.11
            & 0.29/0.60
            & 0.43/0.90
            & 1.32/1.33
            & 2.16/3.65
            & 2.46/5.14 \\

            GSE-Flow~\pub{ICML26} 
            & 0.13/0.24 
            & 0.28/0.59 
            & 0.52/0.99 
            & 0.12/0.24
            & 0.29/0.64
            & 0.53/1.07
            & 0.47/0.89
            & 1.17/2.07
            & 2.66/4.60 \\
            
			\midrule
			\addlinespace[2pt]
			\multicolumn{10}{l}{\textit{General-purpose trajectory forecasting baselines}} \\
			MoFlow~\pub{CVPR25} 
			& 0.07/0.11
			& \underline{0.27}/\underline{0.53} 
			& \underline{0.40}/\underline{0.76} 
			& 0.07/0.11 
			& \underline{0.28}/\underline{0.56} 
			& \underline{0.41}/\textbf{0.81} 
			& 0.44/0.58 
			& 1.04/1.97 
			& 1.85/3.47 \\
            Neuralized MRF~\pub{ICLR25} 
			& 0.09/0.18 
			& 0.68/1.33 
			& 0.90/1.73 
			& 0.10/0.20 
			& 0.73/1.47 
			& 1.07/2.02
			& 0.51/1.01 
			& 2.61/5.40 
			& 4.07/7.98 \\
            PRF~\pub{CVPR26} 
			& \underline{0.04}/0.09
			& 0.29/0.60 
			& 0.42/0.85
			& 0.05/0.10
			& 0.30/0.63
			& 0.49/1.02
			& \underline{0.12}/0.25
			& 0.91/1.98
			& \underline{1.45}/\underline{2.90}\\
            
			\midrule
			\multicolumn{10}{l}{\textit{Physics-guided dynamics}} \\
            
			\textbf{\textsc{PhyODE}} 
			& \textbf{0.03}/\textbf{0.07} 
			& \textbf{0.26}/\textbf{0.51} 
			& \textbf{0.36}/\textbf{0.75} 
			& \textbf{0.04}/\textbf{0.07}
			& \textbf{0.25}/\textbf{0.51}
			& \textbf{0.40}/\underline{0.85}
			& \textbf{0.07}/\textbf{0.13}
			& \textbf{0.48}/\textbf{1.08} 
			& \textbf{0.97}/\textbf{2.00}  \\


			\bottomrule
		\end{tabular}
	}
\end{table*}

\section{Experiments}

\subsection{Settings}

\label{sec: experiments}

\paragraph{Datasets.} We evaluate all methods on the three ExPhy splits. ExPhy-A contains 20k in-distribution scenes, divided into 16k/2k/2k training, validation, and test
sets. ExPhy-B and ExPhy-C each contain 2k held-out test scenes. ExPhy-B
(OOD-Parameter) shifts the object-level physical-property distributions,
whereas ExPhy-C (OOD-Initial) shifts both initial locations and velocities.
All models are trained on the ExPhy-A training set, with checkpoints selected
on its validation set, evaluated on the ExPhy-A test set and ExPhy-B/C without fine-tuning.
We additionally evaluate cross-benchmark transfer on
ComPhy~\cite{chen2025compositional}, an independently constructed video
reasoning benchmark centered on hidden mass and charge. We repurpose its object
trajectories for forecasting and evaluate the ExPhy-trained models without
fine-tuning.

\paragraph{Compared Methods.} 
We compare \textsc{PhyODE} with representative baselines for trajectory
forecasting and physical property estimation. For trajectory forecasting,
\emph{physical reasoning baselines} include VRDP~\cite{vrdp},
PHYCINE~\cite{tang_intrinsic}, and PCR~\cite{chen2025compositional}, whose
visual frontends are replaced with trajectory encoders.
\emph{Geometric dynamics baselines} include PAINET~\cite{yangpainet} and
GSE-Flow~\cite{wuflow}, while \emph{general-purpose forecasting baselines}
include MoFlow~\cite{fu2025moflow}, Neuralized MRF~\cite{NRMF}, and
PRF~\cite{zhou2026recover}. All methods use the same observed trajectories and
prediction horizons. For property estimation, Mean and Uniform Random are input-free baselines.
Mean uses the empirical ExPhy-A training means, while Uniform Random
independently samples mass from $U[0.1,10]$ and friction and restitution from
$U[0,1]$ for each test object. Temporal MLP, Transformer, and Object-GNN are
supervised property predictors trained on ExPhy-A labels. Ground-truth
properties are never provided at inference.

\paragraph{Metrics.}
We evaluate trajectory forecasting using Average Displacement Error (ADE)
and Final Displacement Error (FDE), which measure the average prediction
error over all future steps and the error at the final step, respectively.
Physical property estimation is evaluated using normalized mean absolute
error (NMAE) for mass, friction, and restitution. We divide the corresponding
MAEs by the fixed scales $(r_m,r_\mu,r_e)=(9.9,1,1)$, where $r_m$ is the
ExPhy-A mass span and $r_\mu,r_e$ are the spans of the admissible coefficient
domains $[0,1]$. The same scales are used for all splits, and the average
NMAE is the unweighted mean across the three properties. Lower is better for
all metrics.

\begin{table}[t]
    \centering
    \caption{
    Zero-shot transfer results on ComPhy. All models are trained on ExPhy-A and
    directly evaluated on ComPhy without fine-tuning. We report ADE/FDE
    ($\downarrow$). $\dagger$ indicates trajectory-only adaptations of physical
    reasoning baselines. Best and second-best results are shown in \textbf{bold} and
    \underline{underlined}, respectively.
    }
    \label{tab:comphy_zeroshot}
    \footnotesize
    \setlength{\tabcolsep}{3.8pt}
    \renewcommand{\arraystretch}{1.08}
    \begin{tabular*}{\columnwidth}{@{\extracolsep{\fill}}lccc@{}}
        \toprule
        \multirow{2}{*}{\textbf{Methods}}
        & \multicolumn{3}{c}{\textbf{ComPhy}~\pub{TPAMI25}} \\
         \cmidrule{2-4}
        & Short
        & Mid
        & Long \\
        \midrule

        \multicolumn{4}{@{}l}{\textit{Physical reasoning baselines}} \\
        VRDP$^\dagger$~\pub{NeurIPS21}
        & \underline{0.13}/0.23 & \underline{0.51}/\underline{0.92} & 0.79/1.32 \\
        PHYCINE$^\dagger$~\pub{CVPR23}
        & \underline{0.13}/0.24 & 0.56/0.99 & 0.84/1.41 \\
        PCR$^\dagger$~\pub{TPAMI25}
        & 0.19/0.34 & 0.77/1.41 & 1.15/2.00 \\

        \midrule
        \multicolumn{4}{@{}l}{\textit{Geometric dynamics baselines}} \\
        PAINET~\pub{ICLR26}
        & 0.27/0.36 & 1.70/3.52 & 1.56/2.78 \\
        GSE-Flow~\pub{ICML26}
        & 0.21/0.32 & 0.82/1.18 & 1.42/1.95 \\

        \midrule
        \multicolumn{4}{@{}l}{\textit{General-purpose trajectory forecasting baselines}} \\
        MoFlow~\pub{CVPR25}
        & 0.24/0.37 & 0.89/1.50 & 1.20/1.93 \\
        Neuralized MRF~\pub{ICLR25}
        & \underline{0.13}/\textbf{0.20} & 0.56/0.93 & \underline{0.63}/\underline{1.00} \\
        PRF~\pub{CVPR26}
        & 0.24/0.44 & 1.48/2.70 & 3.09/5.52 \\

        \midrule
        \multicolumn{4}{@{}l}{\textit{Physics-guided dynamics}} \\
        \textbf{\textsc{PhyODE}}
        & \textbf{0.12}/\textbf{0.20} & \textbf{0.37}/\textbf{0.65} & \textbf{0.51}/\textbf{0.82} \\
        \bottomrule
    \end{tabular*}
    \vspace{-5mm}
\end{table}

\paragraph{Implementation Details.} 

All models are implemented in PyTorch and trained and evaluated on a single
NVIDIA RTX 3090 GPU. Unless otherwise specified, baseline models follow their
original implementations and are trained under the same
observation-prediction horizons. We train \textsc{PhyODE} with AdamW using a
learning rate of $1\times10^{-4}$, weight decay of $1\times10^{-5}$, batch
size 64, and 50 epochs.

\begin{table}[t]
    \centering
    \caption{
    Object-level property estimation under the Long setting
    ($T_{\mathrm{obs}}=30$). Entries report NMAE on ExPhy-A/ExPhy-B
    (ID/OOD-Parameter).
    All learned models are trained on ExPhy-A and evaluated zero-shot on ExPhy-B.
    ``Prop. only'', ``Traj. only'', and ``Full'' use
    $\mathcal{L}_{\mathrm{prop}}$, $\mathcal{L}_{\mathrm{traj}}$, and their joint
    objective, respectively. Lower is better; \textbf{bold} denotes the best
    results, including ties.
    }
    \label{tab:property_estimation}
    \renewcommand{\arraystretch}{1.12}
    \setlength{\tabcolsep}{5.0pt}

    \resizebox{\columnwidth}{!}{%
    \begin{tabular}{@{}lcccc@{}}
        \toprule
        \multirow{2}{*}{\textbf{Method}}
        & \textbf{Mass} $\downarrow$
        & \textbf{Fric.} $\downarrow$
        & \textbf{Rest.} $\downarrow$
        & \textbf{Avg.} $\downarrow$ \\
        & \textbf{A/B}
        & \textbf{A/B}
        & \textbf{A/B}
        & \textbf{A/B} \\
        \midrule

        \multicolumn{5}{@{}l}{\textit{Non-learned baselines}} \\
        Mean
        & 0.25/0.77
        & 0.17/0.34
        & 0.22/0.41
        & 0.21/0.51 \\

        Random
        & 0.33/\textbf{0.75}
        & 0.31/0.39
        & 0.30/0.41
        & 0.31/0.52 \\

        \midrule
        \multicolumn{5}{@{}l}{\textit{Supervised property predictors}} \\
        Temporal MLP
        & 0.24/0.77
        & 0.13/0.28
        & 0.17/0.33
        & 0.18/0.46 \\

        Transformer
        & 0.25/0.77
        & 0.11/0.24
        & 0.14/0.27
        & 0.17/0.43 \\

        Object-GNN
        & \textbf{0.22}/0.79
        & \textbf{0.09}/\textbf{0.21}
        & \textbf{0.13}/\textbf{0.27}
        & \textbf{0.15}/\textbf{0.42} \\

        \midrule
        \multicolumn{5}{@{}l}{\textit{\textsc{PhyODE} variants}} \\
        \textsc{PhyODE} (Prop. only)
        & \textbf{0.22}/0.78
        & \textbf{0.09}/\textbf{0.21}
        & \textbf{0.13}/\textbf{0.27}
        & \textbf{0.15}/\textbf{0.42} \\

        \textsc{PhyODE} (Traj. only)
        & 0.30/0.99
        & 0.63/0.63
        & 0.26/0.41
        & 0.40/0.68 \\

        \textsc{PhyODE} (Full)
        & 0.25/\textbf{0.75}
        & 0.17/0.34
        & 0.22/0.40
        & 0.21/0.50 \\

        \bottomrule
    \end{tabular}%
    }
    \vspace{-4mm}
\end{table}

\subsection{Quantitative Results}

\paragraph{Trajectory forecasting.}
Table~\ref{tab:main_results_merged} reports ADE/FDE on ExPhy-A/B/C across
three horizons. \textsc{PhyODE} achieves competitive ID and OOD performance,
including the best Long-horizon result of 0.36/0.75 on ExPhy-A and the best ADE
on ExPhy-B. Its advantage is most pronounced on ExPhy-C, reducing Long-horizon
ADE/FDE from the second-best 1.45/2.90 to 0.97/2.00. Since ExPhy-C shifts both initial locations and velocities while preserving the property ranges, the pronounced gains suggest that the structured dynamics
design of \textsc{PhyODE} is effective for long-horizon extrapolation to unseen
initial states.

\paragraph{Cross-benchmark transfer.}
Table~\ref{tab:comphy_zeroshot} reports zero-shot forecasting results on
ComPhy, where all models are trained only on ExPhy-A and evaluated without
fine-tuning. \textsc{PhyODE} achieves the best performance across all horizons,
with ADE/FDE of 0.12/0.20, 0.37/0.65, and 0.51/0.82 from Short to Long. Its
gains over Neuralized MRF become more pronounced at the Mid and Long horizons.
These results provide external validation that dynamics learned from ExPhy
remain useful beyond its native scene distribution, while
the sustained Mid- and Long-horizon advantages suggest that the structured
dynamics design remains effective under cross-benchmark transfer.

\paragraph{Physical property estimation.}
On ExPhy-A, dedicated property predictors outperform the non-learned
baselines, with Object-GNN and \textsc{PhyODE} (Prop. only) achieving the best
average NMAE of 0.15. On ExPhy-B, errors increase across all methods,
especially for mass, while friction and restitution generalize more reliably;
the same two models remain the strongest overall with an average NMAE of 0.42.
Among the \textsc{PhyODE} variants, property-only supervision achieves
0.15/0.42 on ExPhy-A/B, trajectory-only supervision degrades to 0.40/0.68,
and joint training improves the results to 0.21/0.50 but remains less accurate
than direct property supervision. Together with the forecasting results, these
findings show that low trajectory error does not necessarily imply accurate
physical property estimation.

\begin{figure}[ht]
	\centering
	\includegraphics[width=\linewidth]{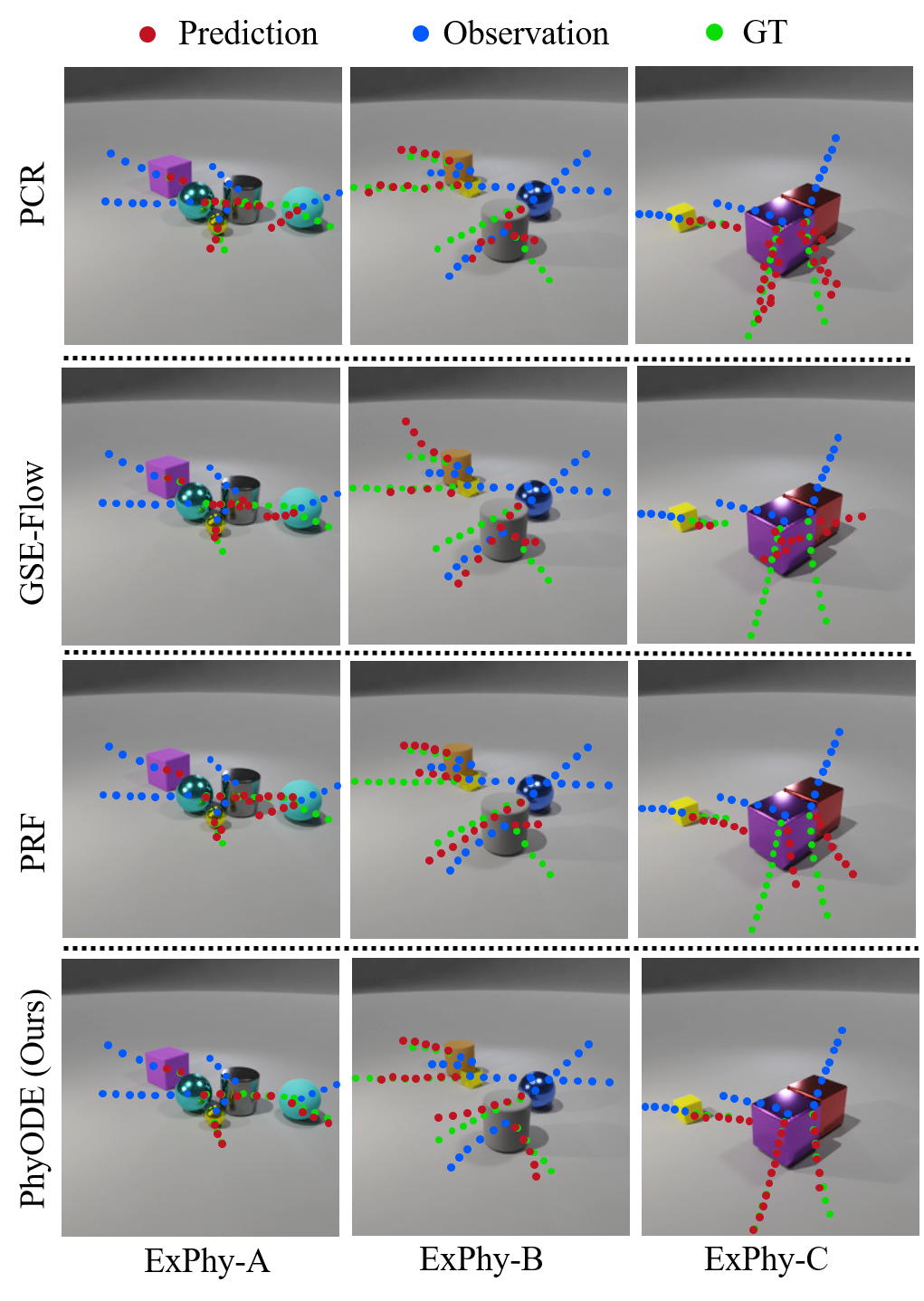}
    \vspace{-5mm}
	\caption{Visualization of long-horizon trajectory forecasting on ExPhy-A, ExPhy-B, and
    ExPhy-C. Rows show representative methods from different model families, and
    columns correspond to ID/OOD splits. Red, blue, and green dots denote predicted,
    observed, and ground-truth trajectories, respectively.}
	\label{fig: compared results}
\end{figure}

\begin{figure}[!ht]
	\centering
	\includegraphics[width=\linewidth]{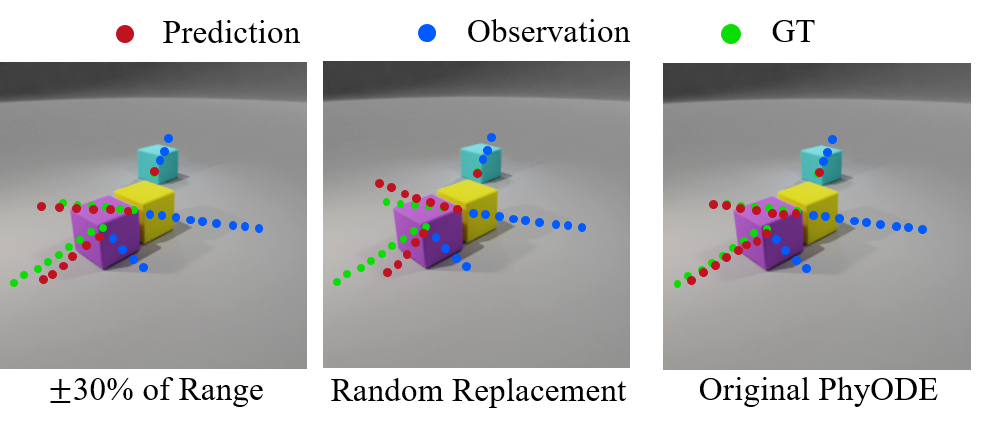}
    \vspace{-7mm}
	\caption{
    Qualitative property interventions under the Long setting. Red, blue, and green denote predicted,
    observed, and ground-truth trajectories, respectively.
    }
	\label{fig:property_intervention}
\end{figure}

\subsection{Qualitative Results}
Figure~\ref{fig: compared results} compares long-horizon forecasts on
ExPhy-A/B/C, while Figure~\ref{fig:property_intervention} qualitatively
examines inference-time property interventions. \textsc{PhyODE} produces
stable predictions under both ID and OOD settings. Compared with the original rollout, fixed perturbations of $\pm30\%$ of each property range and random property replacement produce visible trajectory
changes, providing qualitative evidence that the estimated properties actively influence trajectory rollout.

\begin{table}[t]
    \centering
    \caption{
    Component ablation of \textsc{PhyODE} under the Long horizon setting
    ($T_{\text{obs}}$-$T_{\text{pred}}=30$-$60$). We report
    trajectory forecasting errors as ADE/FDE ($\downarrow$) on both ID and OOD splits. Best results are shown in \textbf{bold}.
    }
    \vspace{-1mm}
    \label{tab:ablation}
    \footnotesize
    \setlength{\tabcolsep}{3.8pt}
    \renewcommand{\arraystretch}{1.10}
    \resizebox{\columnwidth}{!}{
    \begin{tabular}{lccc}
        \toprule
        \multirow{2}{*}{\textbf{Variant}} 
        & \textbf{ExPhy-A} 
        & \textbf{ExPhy-B} 
        & \textbf{ExPhy-C} \\
        \cmidrule(lr){2-2}
        \cmidrule(lr){3-3}
        \cmidrule(lr){4-4}
        & ADE/FDE $\downarrow$
        & ADE/FDE $\downarrow$
        & ADE/FDE $\downarrow$ \\
        \midrule

        w/o explicit physics
        & 0.42/0.86 & 0.48/0.99 & 1.80/3.41 \\


        w/o Neural ODE
        & 0.38/0.79 & 0.41/0.86 & 1.12/2.30 \\

        \midrule
        \textbf{\textsc{PhyODE}}
        & \textbf{0.36}/\textbf{0.75} & \textbf{0.40}/\textbf{0.85} & \textbf{0.97}/\textbf{2.00} \\

        \bottomrule
    \end{tabular}
    }
\end{table}

\subsection{Ablation Study}
Table~\ref{tab:ablation} evaluates the main rollout components of
\textsc{PhyODE}. Removing explicit physics causes the largest degradation,
especially on ExPhy-C, indicating the importance of property-conditioned
physical rollout under OOD initial states. Removing the Neural ODE component
also degrades long-horizon forecasting, suggesting that the learnable residual
dynamics complement the structured physical module.

\section{Conclusion}
In this paper, we presented \emph{ExPhy}, a benchmark for joint trajectory
and physical-property evaluation in multi-object trajectory forecasting. ExPhy contains
24,000 dynamic scenes with trajectories, object-level mass, friction, restitution annotations, and controlled ID/OOD protocols. We also
introduced \textsc{PhyODE}, a physics-guided hybrid model that explicitly
estimates these properties and uses them for differentiable rollout.
Experiments on ExPhy and zero-shot transfer to ComPhy show that low trajectory
error does not necessarily imply accurate physical property estimation.


\bibliography{aaai2027}

\clearpage

\end{document}